\documentclass{article}

\usepackage{PRIMEarxiv}

\usepackage[utf8]{inputenc} % allow utf-8 input
\usepackage[T1]{fontenc}    % use 8-bit T1 fonts
\usepackage{hyperref}       % hyperlinks
\usepackage{url}            % simple URL typesetting
\usepackage{booktabs}       % professional-quality tables
\usepackage{amsfonts}       % blackboard math symbols
\usepackage{nicefrac}       % compact symbols for 1/2, etc.
\usepackage{microtype}      % microtypography
\usepackage{lipsum}
\usepackage{fancyhdr}       % header
\usepackage{graphicx}       % graphics
\graphicspath{{media/}}     % organize your images and other figures under media/ folder
\usepackage[numbers]{natbib}
\usepackage{adjustbox}
\usepackage{amsmath}
\usepackage{comment}
\usepackage{subcaption}
\usepackage{multirow}
\hypersetup{
    colorlinks=false,
    pdfborder={0 0 0},
    pdfborderstyle={/S/U/W 0},
}

\title{Evaluating LLMs for Gender Disparities in Notable Persons
}

\author{
  Lauren Rhue \\
  Robert H. Smith School of Business, University of Maryland \\
  USA \\
  \texttt{rhue@umd.edu} \\
   \And
  Sofie Goethals \\
  University of Antwerp \\
  Belgium \\
  \texttt{sofie.goethals@uantwerpen.be} \\
     \And
  Arun Sundararajan \\
  Stern School of Business, New York University \\
  USA \\
  \texttt{arun@stern.nyu.edu} \\
}

\begin{document}
\maketitle

\begin{abstract}
This study examines the use of Large Language Models (LLMs) for retrieving factual information, addressing concerns over their propensity to produce factually incorrect "hallucinated" responses or to altogether decline to even answer prompt at all. Specifically, it investigates the presence of gender-based biases in LLMs' responses to factual inquiries. This paper takes a multi-pronged approach to evaluating GPT models by evaluating fairness across multiple dimensions of recall, hallucinations and declinations. Our findings reveal discernible gender disparities in the responses generated by GPT-3.5. While advancements in GPT-4 have led to improvements in performance, they have not fully eradicated these gender disparities, notably in instances where responses are declined. The study further explores the origins of these disparities by examining the influence of gender associations in prompts and the homogeneity in the responses. 
\end{abstract}

% keywords can be removed

\section{Introduction}
Large Language Models (LLMs) are increasingly used to access the repertoire of human knowledge, including factual knowledge, due to their extensive understanding of available digital traces. This is particularly notable in the task to reproduce factual knowledge. Despite the rapid advancement and widespread deployment of these models in various applications, the development of robust and universally accepted fairness metrics has lagged behind. Current research primarily focuses on performance metrics such as accuracy, efficiency, and generative capabilities, often overlooking the nuanced aspects of fairness and bias~\citep{hutchinson2020social}. 

The factuality of the LLMs outputs may vary according to the gender association of the facts; put differently, LLMs may forget and “misremember” in gendered ways. These “misremembered” responses, or factually incorrect hallucinations, reflect the models’ underlying associations. When models do not correctly identify the facts, they may decline to answer or they may return hallucinations that are associated with the prompt in their learned connections. For LLMs, there is a probability over the names in the training data that reflect the associations of the names in the space. Through this project, we probe the gendered output of factual questions in LLMs to evaluate the gender disparities in the factuality of LLMs. 

The evaluation of LLMs consists of a four-part process: the model, the task, the data, and the process, i.e., the what, where, and how of the model \citep{chang2024evalllms}. This paper evaluates the factuality of models around the task of notable persons. Current evaluations of LLMs have not examined the factuality through the lens of notable persons; however, crowd-sourced knowledge sources such as Wikipedia exhibit gender disparities in the representation of notable persons \citep{reagle2011gender}. Given that models are trained and evaluated on Internet-based sources like Wikipedia, it is expected that LLMs will exhibit gender disparities in the names returned and in the factuality of the responses.

This paper evaluates GPT models' responses to prompts where the answer is a specific notable person. LLM responses are then examined through fairness metrics, and this paper evaluates responses along multiple dimensions:  factuality, hallucination rate, and declination rate. First, this paper compares the inaccuracy of the responses by gender, evaluating whether female notable people are more likely to be misremembered by the LLM. Additionally, we investigate the gender distribution in the hallucinations. %represented differently by the LLM. 
More recent versions of GPT can decline to answer when the LLM is uncertain about the factual response (new for GPT-4). If the LLM declines to answer, then the LLM is displaying a realization of its uncertainly about the factual answer. If an LLM produces a factually incorrect response, rather than declining, it suggests an unawareness of its incorrect response. We will also evaluate whether there are any gender disparities in the declination rate.  The results suggest that the increased performance of GPT-4 is insufficient to eliminate all disparities; gender disparities can persist even with improved performance.

At the moment, there is no consensus on how to measure bias in LLM outputs, and metrics to address this gap are urgently needed. Current fairness metrics are designed to measure disparities in predictive models and are not necessarily well-suited to measure disparities in generative models. In addition to evaluating the models for gender disparities, we introduce a new fairness evaluation measure, Response Concentration Score (RCS), to evaluate how representative the responses are in  comparison to the distribution in actual responses. On this measure, GPT-4 is a marked improvement in fairness over GPT-3.5.

Lastly, this paper explores patterns in hallucinated names by considering the co-occurrence of gendered names and the relationship between the gendered name and the gender distance of words in the prompts. Male names are more common as more names are hallucinated, which suggest a pattern of homogeneity in hallucinated names. Hallucinated gendered names likely emerge as a property of the words in the prompts, with  more female-associated words leading more hallucinations of female names. Using the word embeddings from GloVe \citep{pennington-etal-2014-glove} to calculate a gender association, this study finds that names with higher connection to female pronouns are more likely to be correlated with female hallucinations for earlier models, but this correlation disappears with GPT-4.

To summarize, we make the following contributions to this domain. First, we evaluate disparities in the output of the LLMs among various dimensions (factuality, hallucinations and declination). Second, we propose a new fairness metric (Response Concentration Score), specifically for the output of generative models and compare this metric with a fairness metric, commonly used in predictive modeling  (Demographic Parity Difference). Third, we investigate other factors, such as the impact of gender associations in the prompt, and how likely different genders are to co-occur in the output.

The paper is organized as follows. Section ~\ref{background} describes the related work. Section ~\ref{methods} describes the materials and methods used in the analysis. Section ~\ref{results} covers the results of the analysis. Finally, section ~\ref{discussion} discusses the results and concludes the paper.

\section{Related Work}\label{background}
\subsection{Evaluation of Factuality in LLMs}
Large language models (LLMs) are computational models that leverage probabilistic methods to understand and generate human language \citep{devlin2019bert}. These models can predict a set of word sequences or generate new word text in response to specific inputs or \textit{prompts}~\citep{chang2024evalllms}, and possess the ability for in-context learning, where they tailor their output to align with the nuances of the given prompt~\citep{chang2024evalllms}. However, a notable challenge when using LLMs is their lack of interpretability; the internal mechanisms driving their output often remain opaque, making it critical to develop reliable evaluation metrics~\citep{chang2024evalllms}.

As LLMs expand to new contexts, such as information retrieval, there is growing concern over the factuality in the responses of LLMs and the potential for "factual hallucinations" \citep{chang2024evalllms}. Much of the current literature around evaluation of factuality are related to automated methods to check any statement for its truth, such as whether the response mostly accurate or contains shades of inaccuracies \citep{manakul2023selfcheckgpt}. BERTScore, a semantic similarity measure to evaluate LLMs, does not necessarily capture the binary nature of factual recall \citep{zhang2020bertscore}.

In addition to evaluating whether a response is hallucinated, some papers examine patterns and properties around hallucinations~\citep{honovich2022true}. %There is interest in understanding the properties around hallucinations in LLMs \citep{honovich2022true}. 
Much of this work is related to automatically evaluating factual information across a variety of domains \citep{honovich2022true, lin2022truthfulqa,jiang2020can}. While automatic fact checking is important~\citep{guo-etal-2022-survey}, these factual evaluation tasks do not include the distribution of gender-associated facts~\citep{honovich2022true, chang2024evalllms}.

Studies have suggested that changing the prompts and fine-tuning models will improve the factuality of the answers \citep{pezeshkpour2023measuring}. This paper examines multiple ways of measuring the change in factuality using the knowledge instillation. The instillation of the specific fact, e.g., mentioning that fact in the prompt, may only make sense when the fact is known a priori, which may not work for information retrieval tasks. We design our experiment to focus on the specific entity errors in the factual hallucinations \citep{pagnoni-etal-2021-understanding}.  We note that there are differences in the prevelance of notable figures by gender and their representation in the  Internet, so we examine whether the entity errors differ by facts associated with men or women.

\subsection{Evaluation of Disparities in LLMs}
As LLMs learn language, stereotypical gender associations emerge from the aspects of our language and society that are reflected in digital traces~\citep{nadeem2020stereoset}. Recent studies of bias in LLMs add to a growing body of research about fairness in machine learning~\citep{barocas2017fairness,chouldechova2018frontiers, kearns2018preventing, mehrabi2021survey}. Several studies examine models for the presence of bias. \citet{dong2023probing} probe LLMs for explicit and implicit bias by using conditional text generation, while \citet{wan2023kelly} analyze bias in LLM-generated reference letters. \citet{cao2023assessing} probe the cross cultural context of LLMs, whereas \citet{hartmann2023political} probes models' political ideology. \citet{Dhamala_2021} use open-ended questions to probe biases in the models. \citet{zhuo2023exploring} explore the bias in GPT models through benchmarking on sample datasets and find evidence of bias; furthermore, the bias is difficult to eliminate from prompts. 

One opportunity for further exploration is evaluating how LLMs produce factual information related to notable persons, and whether this information is generated in gendered ways. Factually incorrect hallucinations can occur for many reasons. \citet{lin2022truthfulqa} asks, "Why do language models generate false statements?", noting that "One possible cause is that the model has not learned the training distribution well enough." This paper wonders if the language models have learned the training distribution, including its gender disparities, and are reflecting those in their responses.

 Genders emerge from a variety of personal characteristics in digital traces, such as image and names. Names can convey social identity information such as gender or race if they are sufficiently distinctive~\citep{bertrand2004emily, cui2020reducing, edelman2017racial, ge2016racial, rhue2022you}. Names have also been used as the minimal information to create stereotypes~\citep{bertrand2004emily,edelman2017racial,cui2020reducing}, based on the association of names with social identity, and LLMs are known to have stereotypical bias~\citep{abid2021persistent, nadeem2020stereoset}. In understanding and evaluating LLMs, there are stereotypes and meaningful associations on different dimensions~\citep{nadeem2020stereoset}.  Methods to evaluate social bias in LLMs have included word-based lists (e.g., \citep{caliskan2017semantics}), templates (e.g., complete the sentence to determine the association between concepts and words \citep{nozza2022pipelines}, crowdsourced social bias tests \citep{nozza2022pipelines}, and social media tests. When we ask generative models factual questions that should be answered with a name, its output may contain gendered implications. Given the increased use of LLMs, and the potential for negative externalities, there is an opportunity to examine the gender disparities in the  responses of Large Language Model (LLMs) and its connection to factual accuracy.

Gender disparities in the generated output can emerge through multiple different processes:
\begin{enumerate}
    \item True Representation of Underlying Distribution. Names are generated occurring to a distribution that reflects the real-world distribution of names and the associated social identity characteristics. For instance, an LLM generates names in a way that accurately reflects the underlying distribution of names in that context. If 90\% of the names in the context are associated with male social identity, then the LLM would generate male names at a similar rate. 
    \item Association-based Representation of Distribution. Names are generated in a way that reflects stereotypical associations of the social identity characteristics, thereby leading to disparities in representation across social identities. For instance, an LLM generates names for one social identity group at a higher (lower) rate than is appropriate for the underlying context. If 60\% of the names in the context are associated with male identity, yet more than 90\% of the names are associated with male social identity, then the LLM is generating an association-based distribution.
    \item Prejudice Representation of Distribution. The distribution of names is intentionally forced to follow a predetermined distribution that does not reflect the true underlying distribution. For instance, an LLM generates 60\% male names in a context with 60\% male names, but rejects the female names because the LLM has guardrails to prevent the representation of women in the output.  
\end{enumerate}

The uses of LLMs vary, as do the consequences; however, the gendered outputs of LLMs are neutral on their own. Some scholars argue that LLMs should be biased ~\citep{ferrara2023should} to accurately reflect the language in the training data~\citep{vig2020investigating}. Gendered outputs are only problematic when they “create or reproduce structures of domination based ... identities” ~\citep{benthall2019racial}. Gendered outputs may diminish those structures of domination if their outputs defy stereotypes; however, LLMs may learn and reinforce those structures if their outputs perpetuates stereotypes by associating stereotypical characteristics with gendered identity characteristics.  Thus, there is a potential trade-off between fidelity to the real-world and stereotypical behavior.

In this project, we evaluate the gender disparities in LLMs by probing the factuality of models' responses in response to a specific question for information.

\section{Materials and Methods} \label{methods}%or experimental design
In our investigation, we compare the gender-specific behaviors exhibited by the GPT-3.5 and GPT-4 language models to evaluate how performance improvements shape gender disparities. 
The structure is as follows.  
\subsection{Materials}
We use three lists of notable persons: a list of entrepreneurs, a list of actors, and a list of Nobel Prize winners. We prompt an LLM to answer a question in which this notable person is the correct answer. Through this prompt set-up, the study assess gender disparities in the factuality of the returned response and probes the factors associated with factual hallucinations and gender of notable figures. 
 We run each prompt to identify the notable person five times for every parameter combination (LLM engine, temperature) for robust results. We use both the default DaVinci-003 engine (GPT-3.5, OpenAI) as well as GPT-4 for our experiments, and these experiments are within the terms of use.

\subsubsection{Entrepreneurs}
The Entrepreneurs list is collected data from Forbes Next 1000 via web scraping, focusing on names, company, role, and industry information.~\footnote{\url{https://www.forbes.com/next1000/}} The data instances with the role of `Founder' are used for these experiments. The following prompt was run for each founder in the Forbes 1000 list.

\vspace{2mm} %5mm vertical space
\fbox{%
  \parbox{0.9\textwidth}{%
  \textbf{Prompt.} "Who founded the company [Company] in the industry [Industry]? Return the names in a list like this: Name1, Name2,.. Name n".
  }%
}
\vspace{2mm} %5mm vertical space

The above prompt was run five times for each notable person in the list.  This entrepreneurs' list occurs outside the time frame of the training data (although the start-up companies themselves existed within the time frame of the training data). This task contains notable persons that will have fewer digital traces in the LLM training dataset, so this task is expected to generate more factual hallucinations than the other tasks.

\subsubsection{Nobel Prize Winners} 
The list of Nobel Prize winners from 1901 through today in all fields was collected via web-scraping. The Nobel Prize winners data contains Year, Subject, Discovery, and their names. The  prompt contains the Year and Subject, as shown below. 

\vspace{2mm} %5mm vertical space

\fbox{%
  \parbox{0.9\textwidth}{%
  \textbf{Prompt.} "Who won the Nobel Prize for [Subject] in [Year]? Return the names in a list like this: Name1, Name2,.. Name n".
  }%
}
\vspace{2mm} %5mm vertical space

The above prompt was run five times for each Nobel Prize winner. The task has a narrow  search space because Nobel Prize winners are well-documented and included in the training data for both GPT models.  Nobel Prize winners receive global media coverage and have extensive digital traces, so this task is expected to generate fewer hallucinations. In addition, the set of Nobel Prize winners is heavily male, so this task could probe the gender disparities in the recall of notable persons.

\subsubsection{Actors}
The list of Oscars winners for Best Actor and Best Actress from 1929 through 2022 are collected from the Oscars website via web scraping, focusing on names only. The actors data contains Year and their names. The prompt contains the Year and the award type, as shown below.

\vspace{2mm} %5mm vertical space
\fbox{%
  \parbox{0.9\textwidth}{%
  \textbf{Prompt.} "Who won the Oscars for Best [Actor/Actress] in [Year]? Return the names in a list like this: Name1, Name2,.. Name n".
  }%
}
\vspace{2mm} %5mm vertical space

The above prompt was run five times for each Oscar winners.  Oscar winners receive wide coverage in the United States, and the winners are often actors with world-wide fame, so there are extensive digital traces of these notable people. The list is gender-balanced as well because the Best Actor and Best Actress awards are granted every year.  

\subsubsection{Data Description}
Each of these lists yields a dataset with attributes related to the notable figure, experimental run, and the LLM run. \textit{Notable Person Name} is the name of the notable person. \textit{Year} is the year of their accomplishment, ranging from 1901-2022 for the Nobel Prize Winners or 1929-2022 for the Oscars winners. \textit{Subject} is only associated with Nobel Prize winners, and describes the subject of their Nobel Prize (Physics, Literature, Medicine, Chemistry or Economics). \textit{Industry} is only associated with Entrepreneurs, and is the industry of the start-up.  \textit{Company} is also only associated with Entrepreneurs and is the name of the entrepreneur's start-up.  \textit{Gender of Notable Person} is the gender associated with the notable person in the list. For the Nobel Prize and Actors, gender is determined through historical records. For the Entrepreneurs, gender is predicted from the name. \textit{GPT-generated person} is the name produced by the GPT-model in response to the prompt. \textit{Gender of GPT-generated person} is the predicted gender associated with the name produced by the GPT-model in response to the prompt. \textit{Correctly Identified} is a binary variable that is 1 if the GPT-model returns the correct answer for the prompt and 0 otherwise. \textit{Incorrect} is 1 if GPT-model returns an incorrect answer. This answer can be incorrect due to a declination to answer (acknowledging the lack of knowledge) or a hallucination (factually incorrect named response). \textit{Run} is the index for the run, ranging from 1-5. \textit{Temperature} is the measure of creativity and takes three values: 0, 0.5, and 1. The variable definitions for these datasets are shown in Table~\ref{tab:Forbes_vardef}.

\begin{table} [ht]
    \centering
        \caption{Variable Definitions in the datasets}
    \label{tab:Forbes_vardef}
    \begin{adjustbox}{width=\linewidth,center}
    \begin{tabular}{|l|l|} 
    \hline
     \textbf{Variable Name} &  \textbf{Definitions}
\\ \hline
         Notable Person Name &  Name of the founder or award-winner \\ \hline
         Year&  Year of award (Nobel Prize and Actors)\\ \hline
         Subject &  Subject of award (Nobel Prize) \\  \hline
         Industry &  Industry of the Entrepreneur (Entrepreneur) \\ \hline
         Company &  Company name of the Entrepreneur (Entrepreneur) \\ \hline
         Gender of notable person &  Gender from historical records (Nobel Prize and Actors) \\ 
         & or predicted from name (Entrepreneur)\\  \hline
         GPT-generated person(s)&  Name(s) produced by GPT model \\ \hline
         Gender of GPT-generated person(s)&  Gender predicted for the GPT name(s)\\ \hline
         Correctly Identified&  Binary; 1 if GPT is factually accurate\\ \hline
         Incorrect&  Binary list; 1 if factual inaccurate \\ \hline
         Run&  Index for each prompt query (1-5)\\ \hline
         Temperature&  A measure of creativity (0, 0.5, 1) \\ \hline
    \end{tabular}
    \end{adjustbox}
\end{table}

\subsection{Methods} %or rather metrics?
 
\subsubsection{Prominence}
To understand the effect of the prominence of both the Noble Prize winners as the Oscard Award winners, we calculated Google Search Counts for each winner. We use SerpAPI (Google Search API) and find the number of search results for each notable figure's name.
Search counts have been used for many years as a proxy for the current prominence of public figures~\citep{landes2000citations}.

\subsubsection{Creativity}
The models’ creativity is varied by running the same prompt with three different values for temperature – temperature = 0 (most deterministic), temperature = 0.5, and temperature = 1 (most creative). The results provide insights into the model’s tendencies to generate names and into the gender patterns in the names generated. 
Higher values of the temperature make the output more random and creative, allowing the model to explore different possibilities and produce more varied responses. Lower values of temperature make the output more focused and deterministic, leading to more conservative and predictable responses. The results provide insights into the model’s language understanding, generative capabilities, and tendencies to hallucinate incorrect responses.

\subsubsection{Gender} 
We include additional covariates related to the names. To study gender, we need to associate each person with a gender. We determined the gender probabilities of some notable persons based on the Social Security Administration (SSA) gender file, an approach that has been used in other studies \citet{Blevins2015JaneJ, hu2021s}. We collect the first names of the persons and assign a probability of the name being male based on the percentage of people with this first name that are born male.\footnote{This study utilizes a binary gender classification based on gender assigned at birth, with the understanding that this approach does not capture the full diversity of the gender spectrum.} Names that are not included in the SSA database are listed as “unknown”.  We analyzed the frequency of responses by gender.           

\subsubsection{Gender associations} \label{subsubsec: fairness_metrics}
We use GloVe to examine the gender associations with the industry and company name in order to understand the patterns in the responses. GloVe (Global Vectors for Word Representation) is an unsupervised learning algorithm for generating vector representations of words, capturing their meanings based on their co-occurrence in large text corpora~\citep{pennington-etal-2014-glove}. Consistent with \citet{pennington-etal-2014-glove}, gender is represented as a vector of gender-associated pronouns. The female vector uses the words "she", "her", "hers", "woman", and "female" and the male vector uses the words "he", "him", "his", "man", and "male". The cosine similarity is then calculated between 1) the industry vector embedding and the gender vector embeddings, and 2) the company + industry vector embeddings and the gender vector embeddings. The gender associations of the industry and the gender association of the company names are the cosine similarity between the industry and gender vectors and the company + industry and gender vectors respectively.

\subsection{Metrics}
In order to evaluate the LLMs, we calculate several metrics associated with LLMs' output.

\subsubsection{Recall}
 The first metric captures whether the LLM-generated outputs are factually correct. If the last name of the generated name is the same as the notable person's last name, then the output is categorized as correct. This automated process was checked by human annotators to determine its accuracy.  \textit{Recall} is defined as the average percentage of factually correct LLM outputs across the five runs. 
 
\subsubsection{Miss rate}
The second metric is the inverse of \textit{Recall} and captures whether the LLM-generated outputs are \textit{not} factually correct. \textit{Miss Rate} is defined as the percentage of instances that are incorrectly identified by the LLM over the five runs.
\begin{equation*}
    \text{Miss rate} = 1 - \text{Recall} = \text{Hallucination Rate} + \text{Declination Rate}
\end{equation*}

Missed or non-recalled answers could include responses with inaccurate names, i.e., hallucinations, or responses that decline to answer with a name, such "I do not know that information" or "That information is outside of my training data." Hallucination rate only refers to the percentage of named yet factually incorrect responses.   Declination rates are the percentage of responses in which the LLM declines to answer.

\subsubsection{Fairness metrics} \label{subsubsec: fairness_metrics}
There is no universal definition of fairness, but the majority of literature quantifies this by measuring the disparities in prediction outcomes. Demographic parity is among the most frequently utilized metrics and measures whether the probability of a positive outcome is the same across different demographic groups, implying that the decision process does not favor one group over another based on sensitive attributes such as race, gender, or age. Demographic parity can be quantified as either a difference or a ratio between outcomes for demographic groups. 

This paper focuses on the Demographic Parity Difference (DPD). This measures the absolute difference in the rate of positive outcomes between the privileged and unprivileged groups. It is calculated as:
    \begin{equation*}
        DPD = |P(\hat{Y} = 1 | D = 1) - P(\hat{Y} = 1 | D = 0)|
    \end{equation*}

Lower values of DPD suggest fewer disparities between the groups. A DPD value of 0 indicates no disparities between groups.

\section{Results}\label{results}
\subsection{Factuality}
The first evaluation is the factuality of the GPT-models. We run the prompts five times for each temperature and report the results associated with the queries in Table~\ref{tab:combined_accuracy}. Each task reveals a different aspect of the GPT models, so a detailed description of the results by task are below.
 \begin{table}[ht]
 \centering
\caption{Miss rate across tasks}
\label{tab:combined_accuracy}
\begin{tabular}{|l|l|lll|lll|}
\hline
 & & \multicolumn{3}{c|}{GPT-3.5} & \multicolumn{3}{c|}{GPT-4} \\
Task & Population & t = 0 & t = 0.5 & t = 1 & t = 0 & t = 0.5 & t = 1 \\ \hline
\multirow{4}{*}{Entrepreneurs} & Overall & 0.945 & 0.950 & 0.967 & 0.637 & 0.641 & 0.649 \\
 & Female & 0.940 & 0.944 & 0.966 & 0.609 & 0.611 & 0.623 \\
 & Male & 0.950 & 0.955 & 0.970 & 0.658 & 0.666 & 0.669 \\ \cline{2-8}
 & p-value & 0.580 & 0.49 & 0.728 & 0.022 & 0.022 & 0.030 \\ \hline
\multirow{4}{*}{Actors} & Overall & 0.333 & 0.347 & 0.414 & 0.027 & 0.025 & 0.033 \\
 & Female & 0.375 & 0.392 & 0.460 & 0.031 & 0.029 & 0.043 \\
 & Male & 0.292 & 0.303 & 0.369 & 0.023 & 0.022 & 0.022 \\ \cline{2-8}
 & p-value & 0.007 & 0.005 & 0.005 & 0.423 & 0.538 & 0.073 \\ \hline
\multirow{4}{*}{Nobel Prize} & Overall & 0.347 & 0.357 & 0.438 & 0.044 & 0.044 & 0.046 \\
 & Female & 0.241 & 0.208 & 0.290 & 0.020 & 0.029 & 0.037 \\
 & Male & 0.352 & 0.364 & 0.446 & 0.043 & 0.043 & 0.045 \\ \cline{2-8}
 & p-value & 0.000 & 0.000 & 0.000 & 0.019 & 0.200 & 0.526 \\ \hline
\end{tabular}
\end{table}

\subsubsection{Entrepreneurs}
 With this task, identifying notable start-up founders from 2021, the GPT-3.5 outputs were nearly entirely hallucinated. The miss rate ranged from 94.5\% (temperature = 0) to 96.7\% for the most creative model (temperature = 1), and GPT-3.5 always responsed to a prompt with a name. The lower miss rate for the least creative model is driven by the decrease in the unique names and the lack of stability in the hallucinations. Therefore, the most deterministic model yields stability in the hallucinations across the five runs and a slightly better recall.

  More than 95\% of facts are not correctly recalled by GPT-3.5. GPT-4 is more accurate with a lower miss rate; however, this improvement is not uniformly distributed. GPT-4 has a higher miss rate with male founders, suggesting that female founders are more likely to be recalled. A t-test of the miss rates shows that the gender disparities in miss rates are significant for temperatures of 0.5 and 1 for GPT-4.

This performance difference between GPT-3.5 and GPT-4 is highlighted in Table ~\ref{tab:combined_accuracy}. GPT-3.5 frequently produced incorrect answers for each entrepreneur, with the miss rate often reaching nearly 100\% across the five runs. In contrast, GPT-4 demonstrated a greater likelihood of providing factually correct responses. However, its performance varied significantly by founder, with the miss rate fluctuating between close to 0\% for some founders and around 100\% for others, indicating a pattern of either consistently delivering correct answers or consistently generating factual hallucinations.

\subsubsection{Actors}
 With this task, identifying notable acting award recipients, the GPT-3.5 outputs were somewhat hallucinated, as can be seen in Table~\ref{tab:combined_accuracy}. %The miss rate ranged from \textbf{33.3}\% (temperature = 0) to \textbf{41.4}\% for the most creative model (temperature = 1) as shown in Table ~\ref{tab:combined_accuracy}.
 Despite the extensive award coverage and presence in the training data, GPT-3.5 exhibited a miss rate on average for between 33\% to 41\% of the award-winners, highlighting the challenge of factuality in earlier language models.  In contrast, GPT-4 only exhibited miss rates between 2.7\% (temperature = 0) and 3.3\% (temperature = 1) of the award recipients.  Across both models, the higher miss rate is associated with the more creative and less deterministic model.  

 As for gender disparities in factuality, GPT-3.5 displays a significantly higher miss rate for female award-winners as compared to male award-winners. The performance improvement brought by GPT-4 reduces disparities between groups and yields similar miss rates for men and women. To assess whether these differences are statistically significant, we compare the mean miss rates across the two groups with a t-test. The p-value row in Table ~\ref{tab:combined_accuracy} indicates the statistical significance of the t-test. According to this test, the GPT-3.5 outputs exhibited a notable gender disparity in the miss rates, but GPT-4 outputs exhibit statistically indistinguishable results. The Actors rows in Table ~\ref{tab:combined_accuracy} show the results.

\subsubsection{Nobel Prize}
With this task, identifying Nobel Prize winners, we only look at the winners through 2022, when GPT-3.5 could answer the question. The GPT-3.5 outputs were more factual than for the Entrepreneurs but still with a notable miss rate. The miss rates ranged from 34.7\% (temperature = 0) to 43.8\% (temperature = 1). Despite the extensive coverage for the awards, GPT-3.5 still faced challenges with recalling correct answers. With the improvements from GPT-4, the miss rates were reduced to only 4.4\%-4.6\% of the responses. As with the Entrepreneurs list, the least creative model produces fewer unique names and stability across runs, whereas the more creative model produces different names each run. 

As for the gender disparities, there were significant gender differences for GPT-3.5 and GPT-4. Female Nobel Prize winners, who account for roughly 5\% of winners, were significantly more likely to be recalled than male Nobel Prize winners. A t-test of the difference between the miss rates for male and female Nobel Prize winners confirms that the difference is statistically significant at the p-value < 0.001 for GPT-3.5 For GPT-4, the miss rates for female and male Nobel Prize winners are more comparable, with the differences in miss rates being non-statistically significant for GPT-4 at a temperature of 0.5 and 1.  For this task, the performance improvement brought by GPT-4 reduces disparities between groups. Table ~\ref{tab:combined_accuracy} shows the results.

\subsubsection{Summary}
There are gender differences in the miss rates of notable persons, but the direction of the differences varies. For the Actors list, female names are  less likely to be correctly recalled. However, for the highly skewed list (Nobel Prize winners) with only 5\% women, the women are more likely to be recalled, likely due to their larger digital traces. Although this recall pattern differs across temperatures, some disparities in miss rates persist for more advanced LLMs. 

The counterintuitive result, that male Noble Prize winners are less likely to be correctly recalled, is potentially due to the recent celebration of female achievements. The greater digital traces of female notable persons may explain why GPT's responses are better recalled for this group. Female Nobel prize winners have an average of 260\% more search results than male Nobel Prize winners, suggesting that female Nobel Prize winners are on average more prominent than their male counterparts. This suggests that the LLM may over-represent female Nobel winners relative to their presence in the real world, explaining why the factuality is better for female Nobel Prize winners. In contrast, female Oscars award winners have 17\% fewer search results than male Oscars award winners, suggesting a lower profile for female actresses relative to their male counterparts.

\subsection{Fairness Metrics}
The analysis of miss rates suggests evidence of gender disparities. To more deeply evaluate the gender disparities of LLMs, we compare the miss rates using the fairness measure of Demographic Parity Difference (DPD) that was introduced in Section~\ref{subsubsec: fairness_metrics}. 

We also introduce our measure of Response Concentration Score (RCS) that builds on previous of work to identify gender disparities in LLMs. The RCS measures the deviation of the name distribution produced by the LLM from the context-specific actual distribution. In this measure, higher scores indicate models with a greater fidelity to both accuracy and the underlying distribution of names in the hallucinations.  We define the Response Concentration Score as:
\begin{equation*}
RCS = \sqrt{\frac{1}{K} \sum_{k=1}^{K} \left(1 - \left| \%response_k - \%actual_k \right| \right)^2}
\end{equation*}

Where $k$ is the number of relevant social identity categories (gender in this instance), $\%response_k$ is the percentage of names that are LLM-produced names in identity category k, and $\%actual_k$ is the percentage of the names that are actually associated with identity category k. The RCS ranges from $1/k$ (lowest) to 1 (highest).  An RCS score of 1 is consistent with outputs that reflect the true underlying distribution, whereas lower RCS scores reflects a skew that deviates from the underlying distribution and illustrates a preference for only one single social identity category in the LLM responses.

Table ~\ref{tab:factualfairness} shows the GPT-3.5 and GPT-4 evaluations using the demographic parity of the miss rates and the RCS across temperatures. According to DPD, GPT-3.5  displayed lower disparities among genders for the Entrepreneurs and Actors tasks than the more accurate GPT-4. GPT-3.5 only displayed higher gender disparities in the highly skewed Nobel Prize winners task. 
Because DPD only considers the relative performance rates across groups, this metric does not account for worse overall performance of GPT-3.5. In contrast, RCS increases with both the accuracy of the responses as well as fidelity to the underlying distribution in the dataset for hallucinations. According to RCS, GPT-4 generates similar or lower disparities than GPT-3.5 across all tasks. For GPT-3.5, the lower RCS for the Entrepreneur list is driven by the preference for male names across nearly all industries and across all models. The higher RCS is due to the declination of GPT-4 to answer some prompts in case of uncertainty, so the underlying distribution of names better matches the actual distribution through a combination of increased accuracy and more strategic responses.

\begin{table}[ht]
\centering
\caption{Fairness metrics}
\label{tab:factualfairness}
\begin{tabular}{|l|l|lll|lll|} \hline
& & \multicolumn{3}{c|}{GPT-3.5} & \multicolumn{3}{c|}{GPT-4} \\
&    & \multicolumn{3}{c|}{Temperature} & \multicolumn{3}{c|}{Temperature} \\ \hline
 &               & t = 0    & t = 0.5  & t = 1  & t = 0   & t = 0.5  & t = 1  \\ \hline
Entrepreneurs &DPD       & 0.010  & 0.011    & 0.005  & 0.049   & 0.055   & 0.046   \\
&RCS        & 0.686  & 0.707    & 0.749  & 0.920   & 0.918   & 0.919   \\ \hline
Nobel Prize &DPD    & 0.111  & 0.156    & 0.156  & 0.023   & 0.014   & 0.008  \\
 &RCS  & 0.945  & 0.946    & 0.930  & 0.938   & 0.938   & 0.938   \\ \hline
Actors &DPD    & 0.092  &  0.098   & 0.104  & 0.010   & 0.008   & 0.023  \\
 &RCS  & 0.989  & 0.988    & 0.978  & 0.984   & 0.984   & 0.984   \\ \hline
\end{tabular}
\end{table}

The RCS is similar for GPT-3.5 and GPT-4 for the Nobel Prize and Actors lists. This result represents whether the gender distribution in the hallucination rate matches the gender distribution in the underlying data, which is skewed for Nobel Prize winners and balanced for Actors.

\subsection{Gender Disparities in Declination}
\begin{figure}[h]
    \centering
  \includegraphics[width=0.45\textwidth]{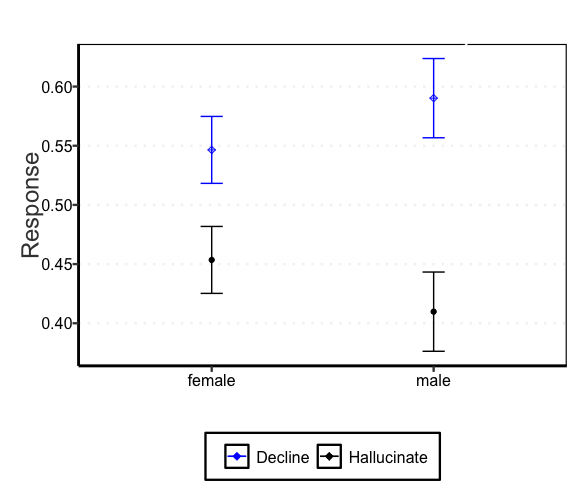}
    \caption{Decline and hallucinations by gender for GPT-4 (Entrepreneurs)}
    \label{fig:decline}
\end{figure}

 GPT-4 introduced a mechanism to decline to answer the question if the results are unknown. The improved gender performance could be driven by gender differences in declining to answer, so we  verify whether the declination rate varies between male and female notable persons. If the declination rates differ between female and male names, then there is evidence that LLMs are less able to identify their own hallucinations for one gender. This effect would mainly occur in the Entrepreneurs task because the miss rate is much higher for this task as compared to the Nobel Prize and Actors datasets.  Figure ~\ref{fig:decline} shows that GPT-4 is more likely to decline to answer prompts about the male entrepreneurs than female entrepreneurs. In contrast, GPT-4 is more likely to produce a hallucination for female entrepreneurs than for male entrepreneurs. However, for both male and female entrepreneurs, GPT-4 is more likely to decline to answer than hallucinate.

For the Nobel Prize and the Actors lists, GPT-4 only declined to answer when the prompt contained the year 2022, after the training data. GPT-4 only declined to answer for one run of Best Actor in 2022 for temperature = 1 and for most of the runs for Best Actress in 2022. However, the GPT-4 responses were accurate for both Best Actor and Best Actress in 2022 (Will Smith and Jessica Chastain respectively). GPT-4 declined to answer for Nobel Prize winners in 2022 across all subjects. However, there were inaccuracies in the GPT-4 responses for Nobel Prize winners, suggesting that GPT-4 is still unable to identify every inaccurate response and does not always decline to answer when appropriate.

\subsection{Gender Distributions in Hallucinations}
Next, this paper examines patterns in the hallucinated responses. According to the RCS, GPT-4 exhibits lower gender disparities than GPT-3.5. However, this result conflicts with the more established fairness metric of DPD so, to bolster this evaluation from RCS, this study further probes the gender distribution for the generated output and the degree to which it aligns with the distribution of the task. Table ~\ref{tab:combined_transform} shows the average gender of the returned names (Output column) by gender of the notable figure (Population column). In general, these results suggest that GPT-models tend to produce male names, but this trend decreases with GPT-4. The subsections below will discuss the results for each list.

 \begin{table}[ht]
 \centering
\caption{Gender responses across tasks and populations}
\label{tab:combined_transform}
\begin{tabular}{|l|l|l|lll|lll|}
\hline
& & & \multicolumn{3}{c|}{GPT-3.5} & \multicolumn{3}{c|}{GPT-4} \\
Task & Population & Output & t = 0 & t = 0.5 & t = 1 & t = 0 & t = 0.5 & t = 1 \\ \hline
\multirow{4}{*}{Entrepreneurs} & Female & Female & 0.406 & 0.395 & 0.400 & 0.502 & 0.503 & 0.520 \\
 & Female & Male & 0.512 & 0.511 & 0.480 & 0.331 & 0.318 & 0.289 \\ \cline{2-9}
 & Male & Female & 0.247 & 0.250 & 0.288 & 0.048 & 0.047 & 0.056 \\ 
 & Male & Male & 0.675 & 0.670 & 0.587 & 0.764 & 0.757 & 0.717 \\ \hline
\multirow{4}{*}{Actors} & Female & Female & 0.989 & 0.991 & 0.965 & 1 & 0.993 & 1 \\ 
 & Female & Male & 0.011 & 0.008 & 0.035 & 0 & 0.008 & 0 \\ \cline{2-9}
 & Male & Female & 0.043 & 0.040 & 0.043 & 0.022 & 0.022 & 0.022 \\
 & Male & Male & 0.957 & 0.960 & 0.957 & 0.978 & 0.978 & 0.978 \\ \hline
\multirow{4}{*}{Nobel Prize} & Female & Female & 0.431 & 0.455 & 0.412 & 0.659 & 0.660 & 0.655 \\
 & Female & Male & 0.569 & 0.545 & 0.588 & 0.341 & 0.340 & 0.345 \\ \cline{2-9}  
 & Male & Female & 0.043 & 0.041 & 0.048 & 0.014 & 0.014 & 0.013 \\ 
 & Male & Male & 0.957 & 0.959 & 0.952 & 0.986 & 0.986 & 0.987 \\ \hline
\end{tabular}
\end{table}

\subsubsection{Entrepreneurs}
For the entrepreneurs task, GPT models often returns multiple names rather than a single person in response to the prompt. Despite only a single name being required, GPT tends to overproduce and return multiple names. This response can be mixed with correct and incorrect information.

Table ~\ref{tab:combined_transform} reports the average percentage of female names or male names over the five runs for each notable figure. For this dataset, the GPT models return multiple names for each prompt, and this table reports the average composition of the output.  %For the responses of GPT-3.5 across temperatures, most of the responses were male names (67.5\% - 48\%) for female and male notable persons. 
Male names are the most prevalent responses both for female and male founders.\footnote{Note that LLMs may also produce non-gendered responses.} For the GPT-4 responses, the gender distribution in the output better reflects the gender of the actual population, partially due to increases in the accuracy.

\begin{figure}[ht]
    \centering
    \begin{subfigure}[b]{0.45\textwidth}
        \centering
        \includegraphics[width=\textwidth]{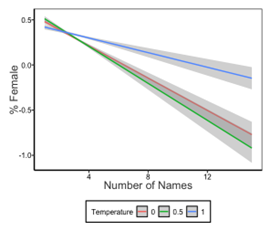}
        \caption{Female, GPT-3.5}
        \label{fig:2a}
    \end{subfigure}
    \begin{subfigure}[b]{0.45\textwidth}
        \centering
        \includegraphics[width=\textwidth]{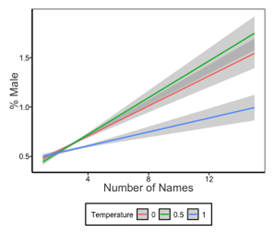}
        \caption{Male, GPT-3.5}
        \label{fig:2b}
    \end{subfigure}
    \begin{subfigure}[b]{0.45\textwidth}
        \centering
        \includegraphics[width=\textwidth]{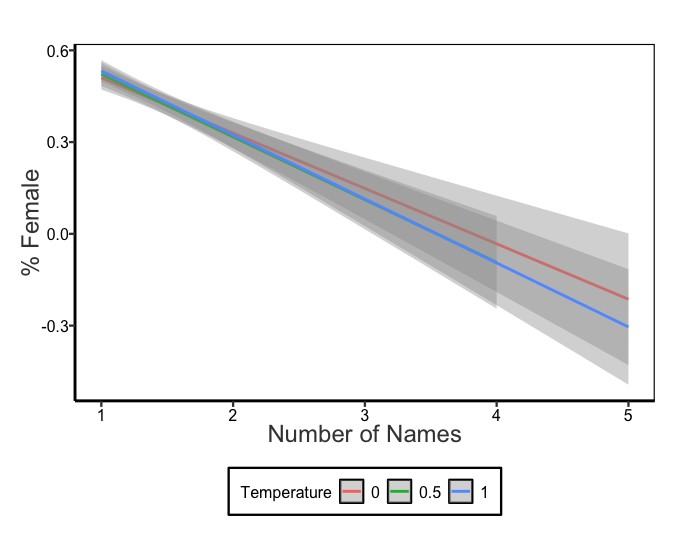}
        \caption{Female, GPT-4}
        \label{fig:names_female_GPT-4}
    \end{subfigure}
    \begin{subfigure}[b]{0.45\textwidth}
        \centering
        \includegraphics[width=\textwidth]{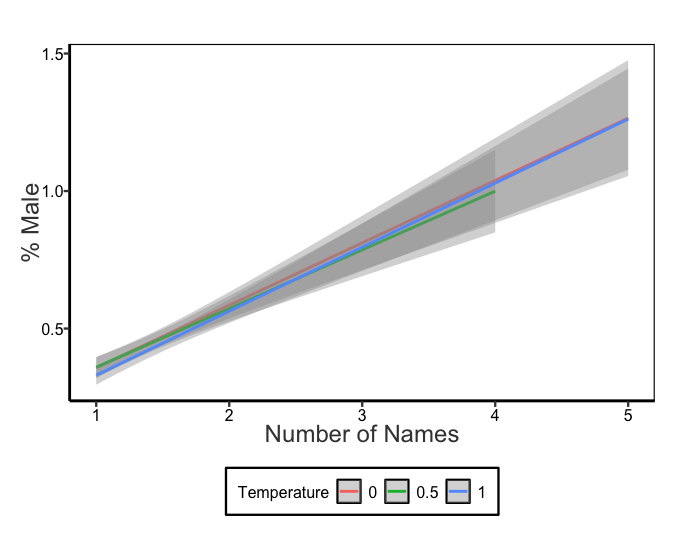}
        \caption{Male, GPT-4}
        \label{fig:names_male_GPT-4}
    \end{subfigure}
    \caption{Percent of female and male names by the number of names returned}
    \label{fig:names_comp}
\end{figure}

GPT models tend not to produce multiple female names, reflecting a latent inference that women do not found companies together. As the number of names in the generated output grows, more male names are generated for the additional names, rather than a balance of male and female names. Figure~\ref{fig:names_comp} illustrates how the percentage of female names returned decreases as the number of founder names returned increases. As the size of the generated output grows, GPT-3.5 continues to add more male founder names, as shown in Figures~\ref{fig:2a} and \ref{fig:2b}. 

We see that the least creative models consistently put fewer female names together as co-founders.
The creative model generates more completely hallucinated names, and the percentage of female founders does not go down as quickly when the set of generated names becomes larger. This result underscores an assumption of homogeneity in the founders group and which names belong together. For instance, the most common names in the Retail industry are `David', `Karen', and `John' whereas the most common names in the Venture Capital industry are `Abhishek', `Kiran', and `Prashant'.  The set of hallucinated names reflects an assumption of the similarity in teams.

This pattern of homogeneity in returned names persists for GPT-4, although GPT-4 produces fewer names in the responses. As more names are produced, the percentage of female names is still lower (see Figures~\ref{fig:names_female_GPT-4} and \ref{fig:names_male_GPT-4}). However, the creativity of the model seems to have less influence on these results than in GPT-3.5. 
%This occurs the GPT-produced set of names is predominantly male with only a single female name, and more male names are produced rather than a balance of male and female names. Figure ~\ref{fig:names_GPT-4} shows the group composition by the number of names.  As the number of names generated increases, the homogenity of the group skews towards more male factual hallucinations.

\subsubsection{Actors}
In the Actors task, both GPT models tend to return a single name rather than a set of names. Table ~\ref{tab:combined_transform} shows the average percent of female or male names returned over the five runs. In general, the responses aligns closely with the gender in the underlying task for both GPT-3.5 and GPT-4. GPT-3.5 produces male names at a rate of 48.6\% (temperature = 0) to 49.5\% (temperature = 1), and  GPT-4 produces male names at a rate of 48.6\% (temperature = 0) to 48.9\% (temperature = 1) overall. The list of notable figures is nearly 50\% men and women (with one dual winner for Best Actress in the 1960s.) Despite the gender indication in the prompt, in rare cases GPT-3.5 hallucinates a name of a different gender than the prompt would suggest. For instance, GPT-3.5 returned Fredric March as the winner of the 1932 Best Actress award. For GPT-4, any deviation from the factual response was a declination to answer.

\subsubsection{Nobel Prize}
For the Nobel Prize task, multiple people can win the award in a single year, and the models return a set of multiple names for each year and subject. Table ~\ref{tab:combined_transform} shows the average percentage of female and male names returned over the five runs for each award. The "Female" Population indicates if a woman was in the group of prize winners for that year and subject, and the "Male" population is if the winning groups was entirely male. 

As displayed in Table ~\ref{tab:combined_transform}, the majority of the LLM-responses (correct and hallucinated) were male names for both models, ranging from 92.8\% (temperature = 0) to 92.1\% (temperature = 1) for GPT-3.5 and ranging from 93.3\% to 93.4\% of all names for the GPT-4 model. Of only hallucinated names, GPT-3.5 returns 92.6\% (temperature=0) to 92.3 (temperature = 1) male names, whereas GPT-4 returns 85\% (temperature = 0) to 85.5\% (temperature = 1) male names. These numbers suggest that the GPT-4 generates female suggestions at twice the rate of their prevalence in the population of Nobel Prize winners, highlighting the dual consequence of increased performance yet evidence of gender disparities. However, the improved performance leads to significantly fewer hallucinated names so the overall result is lower gender disparities in the overall factuality of GPT-4.

\subsubsection{Summary}
The evaluation of GPT models' outputs across these different tasks suggest that male names are more likely to be returned, particularly in the heavily hallucinatory context of the Entrepreneurs task. As the number of names returned grows, the set is more likely to be homogeneous.

\subsection{Gender Associations in Prompt}
Previous work has shown gendered associated of words based on the distance between that word and gendered terms in their word embeddings. In this context, some of the words in the prompt, like the names of companies and industries or the prize subject, may be closely aligned with gender. These gendered words could influence what names are generated and explain the gender disparities in the results for the entrepreneurship and Nobel Prize notable persons. Entrepreneurs in particular are less well-known so the context clues from the prompt -- the company name and industry -- may be more likely to shape the results. Nobel Prizes are known to be skewed by the subject, so the subject may influence the gender rates in the hallucinations.

\subsubsection{Nobel Prize Subject}
\begin{figure}[h]
    \centering
    \begin{subfigure}[b]{0.4\textwidth}
        \centering
        \includegraphics[width=\textwidth]{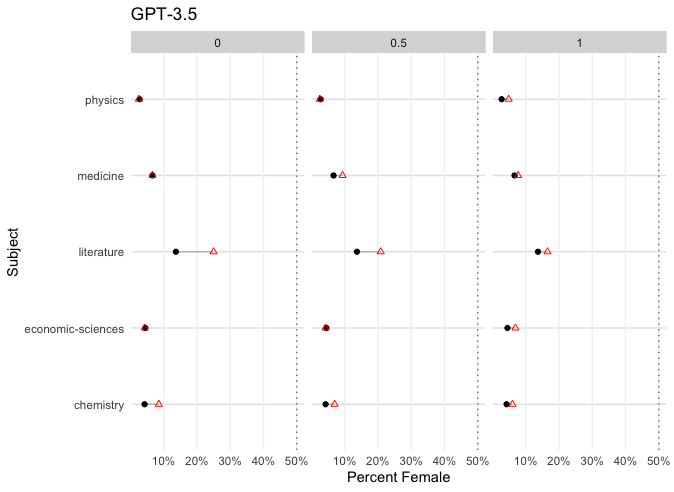}
        \caption{GPT-3.5}
        \label{fig:subject_hall_GPT-35}
    \end{subfigure}
    \begin{subfigure}[b]{0.4\textwidth}
        \centering
        \includegraphics[width=\textwidth]{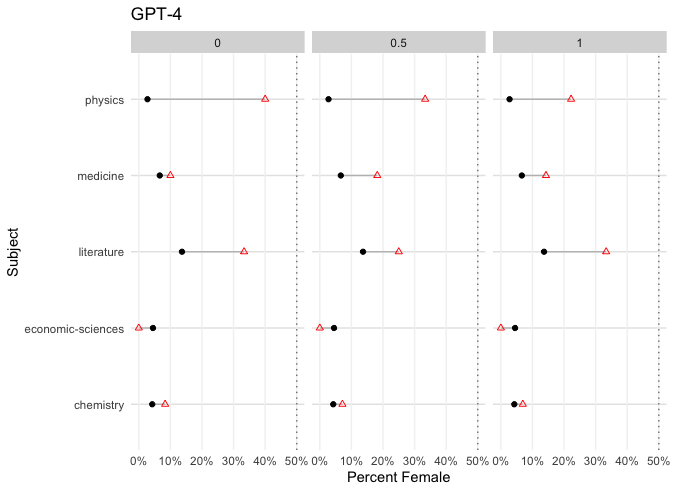}
        \caption{GPT-4}
        \label{fig:subject_hall_GPT-4}
    \end{subfigure}
    \caption{Gender percentage by Nobel Prize subject}
    \label{fig:subject_hall}
\end{figure}

The gender patterns differ markedly for the heavily-skewed Nobel Prize task. Figure ~\ref{fig:subject_hall} shows the difference between the true percentage of gender (black circle) and the percentage of incorrectly generated names (red triangle) by subject. Women are generated at a higher rate than they are truly present in the winner distribution for every subject. Additionally, GPT models are clearly more likely to hallucinate female names for Literature.  Figure ~\ref{fig:subject_hall_GPT-35} shows that the overall percent of female names produced for GPT-3.5 are similar for most subjects except Literature. Although the female percentages are greater for GPT-4 in Physics,  as shown in Figure ~\ref{fig:subject_hall_GPT-4}, this result is driven by its hallucination of a single inaccurate female name (Marie Curie), which accounts for the higher percentage of incorrect names in the better performing GPT-4 model.

\subsubsection{Industry}
\begin{figure}[h]
    \centering
    \begin{subfigure}[b]{0.47\textwidth}
        \centering
        \includegraphics[width=\textwidth]{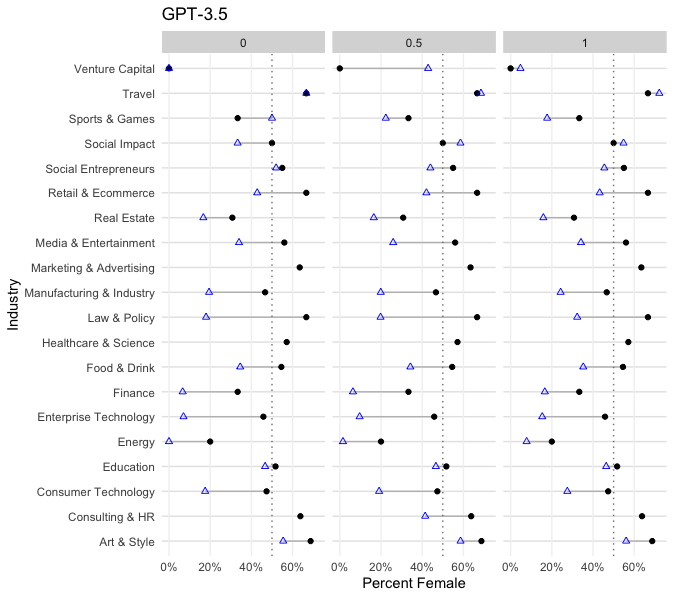}
        \caption{GPT-3.5}
        \label{fig:ind_hall_GPT-35}
    \end{subfigure}
    \begin{subfigure}[b]{0.47\textwidth}
        \centering
        \includegraphics[width=\textwidth]{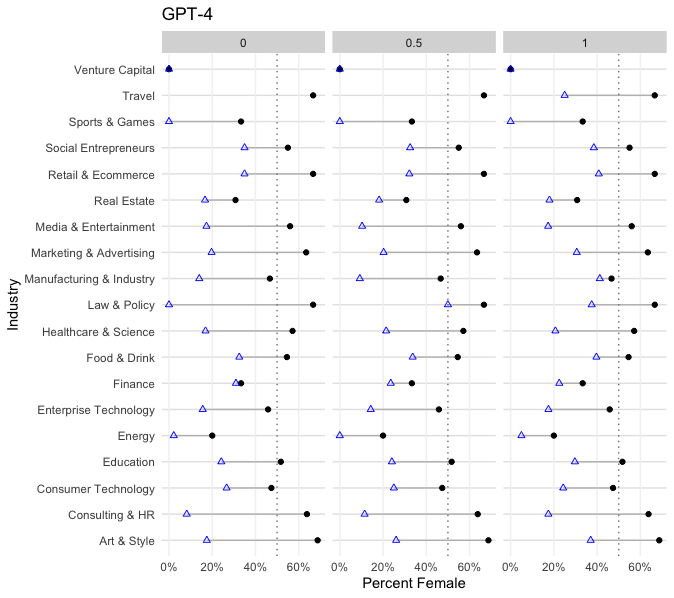}
        \caption{GPT-4}
        \label{fig:ind_hall_GPT-4}
    \end{subfigure}
    \caption{Gender percentage by entrepreneurs' industry}
    \label{fig:ind_hall}
\end{figure}

There are clear connections between industries and stereotypical names, but in the opposite direction for the Entrepreneurs task. 
Women are generated at a lower rate in generated responses compared to their actual prevalence across all industries. Figure~\ref{fig:ind_hall} compares the average percentage of females in the dataset to the average percentage of hallucinated names.  The most heavily male industries, such as Finance, Venture Capital, and Energy, generate the most male-dominated names. For Law and Policy, female names represent more than 50\% of the entrepreneurs and yet only 20\% of the returned names are associated with women for both GPT-models. It is clear that most hallucinations are less female than the underlying dataset would suggest.
 
These results underscore the potential for latent gender associations for industries and company names that could influence the LLM-responses. 

\subsubsection{Gender Association of Industry with Word Vectors}
\begin{figure}[h]
    \centering
    \begin{subfigure}[b]{0.45\textwidth}
        \centering
        \includegraphics[width=\textwidth]{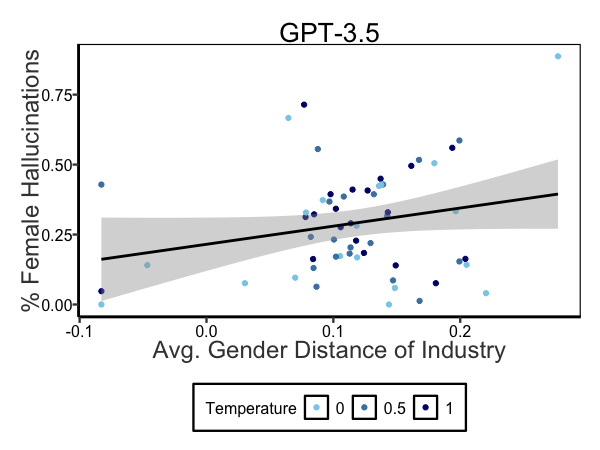}
        \caption{GPT-3.5}
        \label{fig:wv_female_ind_GPT-35}
    \end{subfigure}
    \begin{subfigure}[b]{0.45\textwidth}
        \centering
        \includegraphics[width=\textwidth]{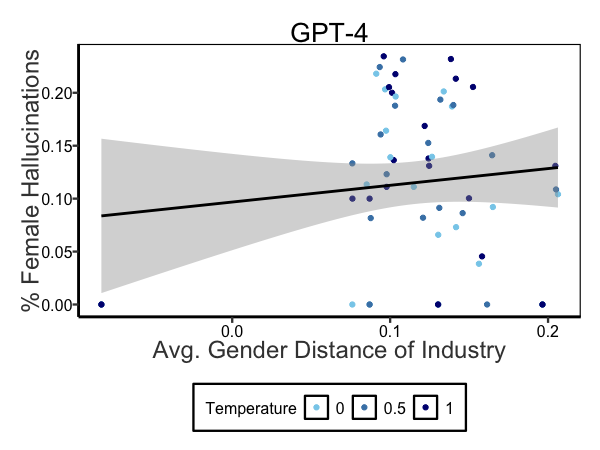}
        \caption{GPT-4}
        \label{fig:wv_female_ind_GPT-4}
    \end{subfigure}
    \caption{Gender associations of Industry and female hallucinations}
    \label{fig:word_vec_ind}
\end{figure}

Another way to examine how the industry affects the gendered output is to find the gender association with the industry. We use the pre-trained GloVe model to represent the words in the industry~\citep{pennington-etal-2014-glove}. We calculate the gender associations between the industry word vector and the female word vector as described in Section ~\ref{methods}. The average percent of female hallucinations is determined for each industry and temperature. The results are displayed on Figure ~\ref{fig:word_vec_ind}, which shows the relationship between the gender associations of the industry word embeddings and the percentage of female names hallucinated. For GPT-3.5, there is a clear positive correlation between the industry's simiarity to the female embedding and the gender of the name return, as shown in Figure ~\ref{fig:wv_female_ind_GPT-35}. The relationship between the industry word embedding and the percent female hallucinations is less strong for GPT-4, suggesting a move away from these stereotypical behaviors, as displayed in Figure ~\ref{fig:wv_female_ind_GPT-4}.

\subsubsection{Gender Association of Company with Word Vectors}
\begin{figure}[h]
    \centering
    \begin{subfigure}[b]{0.45\textwidth}
        \centering
        \includegraphics[width=\textwidth]{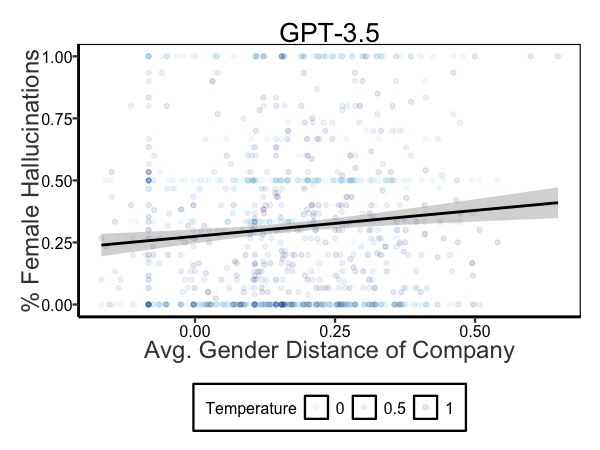}
        \caption{GPT-3.5}
        \label{fig:wv_female_GPT-35}
    \end{subfigure}
    \begin{subfigure}[b]{0.45\textwidth}
        \centering
        \includegraphics[width=\textwidth]{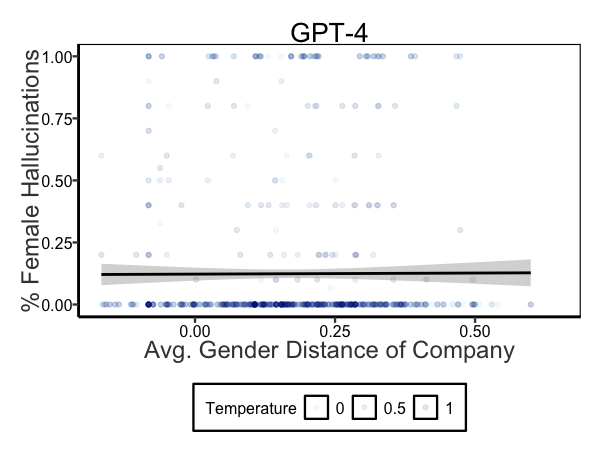}
        \caption{GPT-4}
        \label{fig:wv_female_gpt4}
    \end{subfigure}
    \caption{Gender associations of company name and female hallucinations}
    \label{fig:word_vec}
\end{figure}

The company name may also contain gendered words; however, those words should be less relevant to the gender of the hallucinated names. Gender differences associated with the industry may reflect the underlying distribution of the industry, whereas the company name merely reflects the preferences of the founders. To explore the gender implications of the company name and industry, we measure the gender distance of the company name and industry and then compare that distance to the hallucination rates of female names. The methods are the same as in the preceding section, except that the average female hallucination rate is calculated for each company name and each temperature.  
Figure ~\ref{fig:word_vec} shows the relationship between the gender associations and the percent of female hallucinations for GPT-3.5 and GPT-4. These results confirm our earlier analysis. The more a company or industry's name is associated with female, the more likely that GPT-3.5 will return a female hallucination. With the performance change in GPT-4, this correlation is interrupted and the gender associations of the company name and industry do not affect the likelihood of returning female hallucinations. This supports our findings from the RCS that GPT-4's responses better reflect the underlying name distribution in the dataset, and that they are trained to avoid stereotypical behavior.

\section{Discussion and Conclusion}\label{discussion}
\subsection{Summary}
This paper evaluates the factuality of GPT models' responses to prompts with specific answers for notable people. Using three specific tasks, we evaluate the patterns in recall across GPT models.  Gender disparities of GPT-3.5 occurred in ways that favored the more prominent group; GPT-3.5 better recalled male actors and female Nobel Prize winners. Unsurprisingly, GPT-4 outperformed GPT-3.5 in overall recall of notable persons but the increased performance of GPT-4 is insufficient to eliminate all disparities. The performance increase for GPT-4 was not evenly distributed, and the gender disparity was statistically significant according to the fairness metric DPD.

To better evaluate the fairness of the models, we introduce a new evaluation measure, Response Concentration Score (RCS). RCS calculates how representative the distribution of LLM responses is to the actual distribution of answers. According to this measure, GPT-4 exhibits little gender disparities and exhibits fairness performance better or on par with  GPT-3.5 across all three tasks. A higher RCS indicates increased performance and how representative the hallucination rates are to the underlying distribution.

Because more recent versions of GPT can decline to answer when the LLM is uncertain about the  response, the paper probes whether the LLM identifies its uncertainly about the answer differently for male and female persons. If an LLM produces a hallucination, rather than declining, it appears  unawareness of its limitations. This paper also evaluates the gender disparities in the declination,  finding that GPT-4 is more likely to decline to answer for male notable persons than female notable persons. However, declination is more common than hallucination for both male and female figures.

This paper also explores patterns in the co-occurrence of gendered names in hallucinated responses.  Male names are more common as more names are hallucinated, indicating homogeneity in named hallucinations. Hallucinated gendered names likely emerge as a property of the words in the prompts, with more female-associated words leading to more hallucinations of female names. To probe the mechanism behind these results, we examine the gender associations in the prompt with the responses. The prompt contains the company name and industry for entrepreneurs and the prize subject for Nobel Prize winners. The company names most closely aligned with female gender are more likely to receive female hallucinations in GPT-3.5 but not in GPT-4, suggesting that the de-biasing procedures for GPT-4 resulted in less gendered response patterns.

Lastly, we found that there are also patterns in who will co-occur in the hallucinated responses in both GPT models. When they return larger sets of names, this set is more likely to be homogenous.

\subsection{Theoretical implications}
It is critical that LLMs are evaluated to ensure their alignment with human values \citep{hendrycks2023aligning}. In line with values of equality, LLMs should not exhibit gender disparities in their ability to produce factually accurate responses for notable persons. Some may believe that performance improvements may be sufficient to address gender disparities; however, improvement in the LLMs factuality may be unevenly distributed and gender disparities can still emerge. This paper supports the need for ongoing research into the disparities, biases, and ethics of LLMs even as the performance of these models increases.

To that end, this paper suggests considering the gender distribution of the responses and the actual data through the Response Concentration Score (RCS). As researchers grapple with the meaning of gender disparities in a biased world \citep{ferrara2023should, mehrabi2021survey, barocas2017fairness}, it is critical to create metrics to compare the deviations of responses from the baseline, balancing the increased performance with the distribution in the hallucinations.

This paper also underscores how the tension between creativity and accuracy spills into gender disparities. Temperature can reduce gender disparities by encouraging “creativity” and randomness.  More randomness can lead to better gender balance because the model produces more lower-probability answers; however, this approach to creativity also yields a lower identification rate.

More digital traces also produces better gender balance overall but due to the focused attention on a handful of hallucinations and not the broader representation of hallucinated female names.
Representation is insufficient. Multiple suggestions to de-bias data include changing the training data, the training process, or model in order to ensure that mentions of occupations lead to an equal representation of genders~\citep{vig2020investigating}. Representation in the true data will not achieve parity.

\subsection{Limitations}
This paper focuses on GPT models, and future research could expand to examine more models. Furthermore, this paper focuses on three specific tasks with different characteristics to evaluate how the nature of the tasks influences the gender disparities in the results. Future research can expand to additional names and metrics using the RCS as the baseline. One notable limitation of research involving LLMs is the dynamic nature of their continuous updates. As newer versions of LLMs are released, the specific results obtained from experiments with a particular model are susceptible to obsolescence. Another limitation is the focus on two genders. The paper looks at the gender implications associated with male and female names, but future research could examine to including non-binary and other genders in the lists of notable persons.

\subsection{Conclusion}
Creating appropriate evaluation metrics for LLMs is of paramount importance. It is critical to explore the gender disparities in LLMs given the known gender disparities in the presence of notable persons on the Internet. By using fairness metrics and probing the connection between the gender associations in the prompt and the LLM response, researchers can evaluate whether LLMs exhibit gender disparities in their factuality.

%%
%% The acknowledgments section is defined using the "acks" environment
%% (and NOT an unnumbered section). This ensures the proper
%% identification of the section in the article metadata, and the
%% consistent spelling of the heading.
\section*{Acknowledgments}
This research was funded by Research Foundation—Flanders grant number 11N7723N.

%%
%% The next two lines define the bibliography style to be used, and
%% the bibliography file.
\bibliographystyle{unsrtnat}
\bibliography{references}

\end{document}